\documentclass[letterpaper, 10 pt, conference]{ieeeconf}

\usepackage{tikz}
\usepackage{lipsum}  
\usepackage{atbegshi}
\usepackage{fancyhdr}

\makeatletter
\let\oldmaketitle\maketitle
\renewcommand{\maketitle}{
  \oldmaketitle
  \thispagestyle{fancy}
}
\makeatother

\AtBeginShipout{%
  \AtBeginShipoutAddToBox{%
    \begin{tikzpicture}[remember picture, overlay]
        \node[anchor=south, font=\bfseries\small, text=black] at ([yshift=1.2cm]current page.south) {\thepage};
        \ifnum\value{page}=5515
          \node[anchor=north west, font=\bfseries\small, text=black] at ([xshift=1.8cm, yshift=-0.8cm]current page.north west) {2023 IEEE International Conference on Robotics and Automation (ICRA 2023)};
          \node[anchor=north west, font=\bfseries\small, text=black] at ([xshift=1.8cm, yshift=-1.2cm]current page.north west) {May 29 - June 2, 2023. London, UK};
          \node[anchor=south west, font=\bfseries\small, text=black] at ([xshift=1.8cm, yshift=1cm]current page.south west) {979-8-3503-2365-8/23/\$31.00 ©2023 IEEE};
        \fi
    \end{tikzpicture}%
  }%
}

\IEEEoverridecommandlockouts
\usepackage[utf8]{inputenc} 
\usepackage[T1]{fontenc}    
\usepackage{hyperref}       
\usepackage{url}            
\usepackage{booktabs}       
\usepackage{amsfonts}       
\usepackage{nicefrac}       
\usepackage{microtype}      
\usepackage{xcolor}         

\usepackage{times}
\usepackage{soul}
\usepackage[utf8]{inputenc}
\usepackage[small]{caption}
\usepackage{graphicx}
\usepackage{amsmath}
\usepackage{booktabs}
\usepackage{algorithm}
\usepackage{algorithmic}
\usepackage{ amssymb }
\usepackage{ dsfont }
\usepackage{subcaption}
\usepackage{multirow}
\usepackage{subcaption}
\usepackage{array,etoolbox}
\preto\tabular{\setcounter{magicrownumbers}{0}}
\newcounter{magicrownumbers}

\title{Explainable Action Advising for Multi-Agent Reinforcement Learning}

\author{Yue Guo$^1$, Joseph Campbell$^1$, Simon Stepputtis$^1$, Ruiyu Li$^1$, \\ Dana Hughes$^1$, Fei Fang$^1$, Katia Sycara$^1$, \textit{Fellow, IEEE}
\thanks{$^{1}$Carnegie Mellon University, USA {\texttt{ \{yueguo, feif, katia\} @cs.cmu.edu, \{jcampbell, stepputtis \}@cmu.edu, \{ruiyul, danahugh\}@andrew.cmu.edu}} }%
 \thanks{*This work was supported by DARPA award HR001120C0036 and by AFRL/AFOSR award FA9550-18-1-0251 and AFOSR FA9550-18-1-0097 and ARL W911NF-19-2-0146. Co-author Fang is supported in part by IIS-2046640 (CAREER).}
}

\begin{document}

\maketitle

\begin{abstract}
Action advising is a knowledge transfer technique for reinforcement learning based on the teacher-student paradigm.
An expert teacher provides advice to a student during training in order to improve the student's sample efficiency and policy performance.
Such advice is commonly given in the form of state-action pairs.
However, it makes it difficult for the student to reason with and apply to novel states.
We introduce Explainable Action Advising, in which the teacher provides action advice as well as associated explanations indicating why the action was chosen.
This allows the student to self-reflect on what it has learned, enabling advice generalization and leading to improved sample efficiency and learning performance -- even in environments where the teacher is sub-optimal.
We empirically show that our framework is effective in both single-agent and multi-agent scenarios, yielding improved policy returns and convergence rates when compared to state-of-the-art methods.

\end{abstract}

\section{Introduction}


As robots are deployed in real-world settings, opportunities may arise for agents to impart their knowledge to other agents.
When considering the space of all possible observations that an agent might encounter in the wild, it is unlikely a single agent can learn to react to all possible situations.
In such cases, it is beneficial for robots with a diversity of experiences to share this knowledge with others so they know how to act.
Recent works have explored teacher-student methods~\cite{da2020agents},  \cite{torrey2013teaching}, \cite{gf:Lixto}, \cite{zhu2020learning} in which a teacher agent transfers its knowledge to a student agent with the goal of accelerating the student's learning.
Such techniques can be applied in robotics, where experienced agents can transfer knowledge on how to handle as-of-yet unseen situations to inexperienced students.
Teacher-student methods are applicable in many scenarios, such as when a) directly copying the teacher's knowledge is infeasible, e.g., different policy models or a human student/teacher; b) the teacher is non-optimal and we wish to only transfer a subset of its knowledge, e.g., transfer learning to novel environments; or c) we want to pursue incremental learning as in human curricula.
A well-studied approach is that of \emph{action advising} in which a teacher selectively provides advice in the form of state-action pairs to the student, representing a suggested course of action.
Reinforcement learning (RL) settings in particular benefit from advising, given the high sample complexity associated with many algorithms~\cite{da2020uncertainty} which leads to time-consuming training.
Unlike related techniques such as inverse reinforcement learning~\cite{ho2016generative},
intention inference~\cite{grover2022semantically},
and offline reinforcement learning~\cite{gao2018reinforcement} which assume an offline dataset of expert actions, action advising techniques leverage an intelligent teacher capable of deciding when and how to issue advice.

\begin{figure}
    \centering
    \includegraphics[width=1\linewidth]{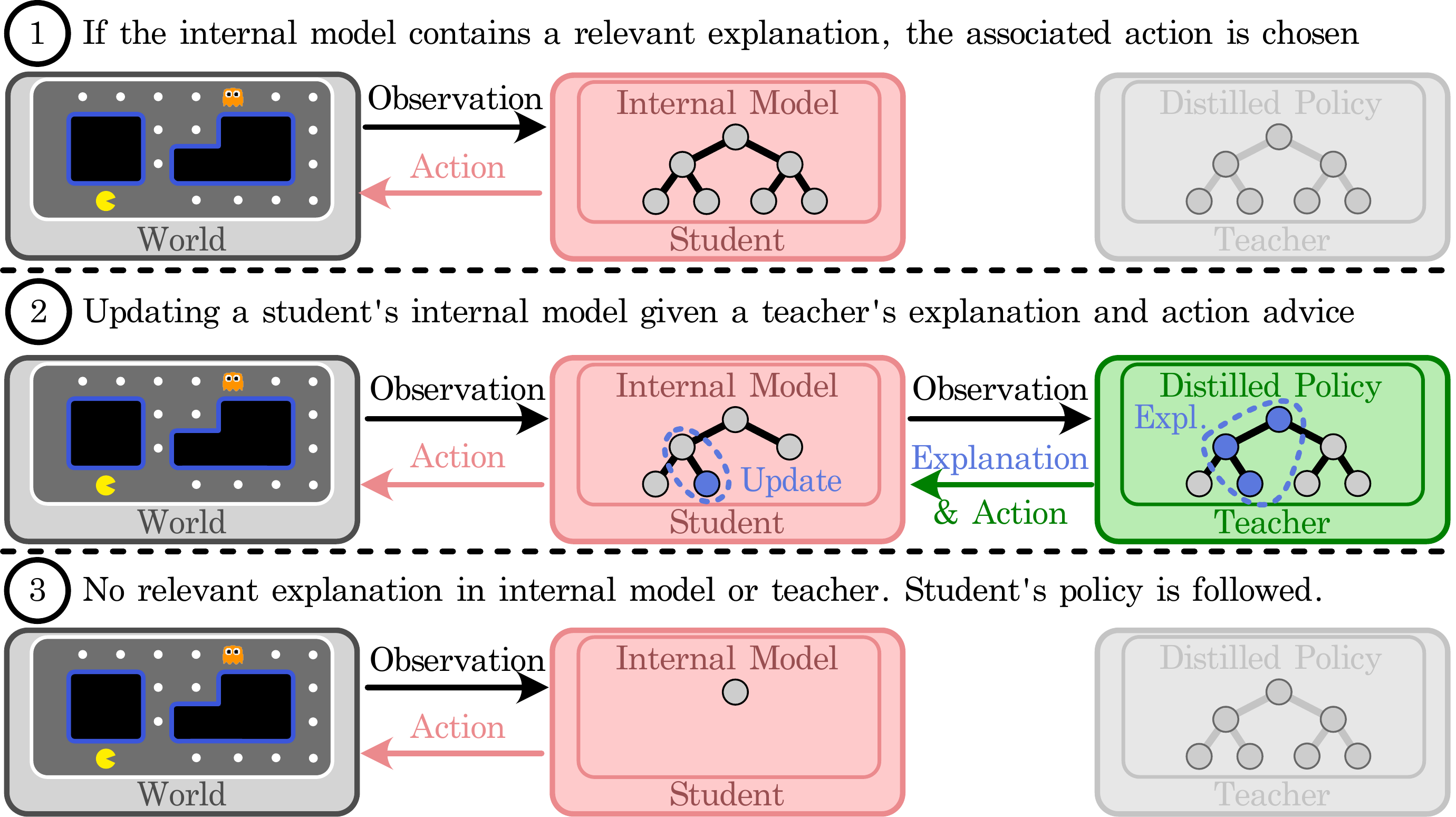}
    \caption{Overview of EAA. As the student makes observations during training, one of three cases may occur. 1) The student has previously received advice for a similar state and is able to identify it through an explanation in its internal model, then follow the associated action. 2) The student has not previously received advice, but is now given advice by the teacher and stores the explanation in its internal model, then follows the advised action. 3) The student has not previously received advice and the teacher is not currently providing advice; the student follows its own policy. In cases where the teacher is trained in a different environment than the student, the student may choose to reject unhelpful advice.
    \vspace{-0.5cm}}
    \label{fig:overview}
\end{figure}

Despite some notable successes in simple environments~\cite{da2020uncertainty}, action advising methods often perform poorly in complex environments with large state spaces.
This is especially true for multi-agent reinforcement learning (MARL) where environments exhibit complex team dynamics, non-stationarity, and high-dimensional state spaces.
Memory-based approaches address these difficulties to a certain extent by reusing previous advice~\cite{zhu2020learning}; however, they are inherently limited as advice can only be re-used when the exact same state is encountered again.
To overcome this, we propose a novel action advising algorithm in which the teacher agent issues action advice as well as an explanation for \textit{why} that advice was issued.
We show that incrementally incorporating these explanations into an internal model enables a form of \emph{learning through reflection}~\cite{helyer2015learning}, where an agent is able to reason about why a specific action was recommended and when it would apply in the future.
This greatly improves advice generalization
and
improved sample efficiency.
Reflection also allows an agent to identify unhelpful advice under certain conditions, allowing advice transfer to novel environments.


In our proposed method, Explainable Action Advising (EAA), we assume access to a teacher's 
``black box''
policy, i.e., neural network, from which we can sample demonstration trajectories.
These trajectories are used to distill the policy into a decision tree in which each leaf represents an action, and the corresponding path from root to leaf represents the explanation for why an action was taken.
When the teacher issues advice, both the decision path (explanation) as well as the leaf (action) are given to the student, who incrementally re-assembles the decision paths together in an attempt to reconstruct the teacher's decision tree in its internal model.
This internal model serves two purposes:
1) it acts as a memory and allows a student to apply prior advice to same or similar future states, and
2) it allows a student to identify unhelpful advice and selectively ignore it, which is useful for a sub-optimal teacher and/or novel environments.
This is analogous to the self-explanations~\cite{chi1994eliciting} employed by human students in order to improve learning effectiveness.
Importantly, EAA is agnostic to the student's underlying reinforcement learning algorithm -- intuitively the teacher is guiding policy exploration.

Our work makes three main contributions, we: 
a) introduce an RL algorithm-agnostic action advising method which uses explanations to improve advice generalization; %
b) utilize explanations to make use of sub-optimal advice for use in transfer learning; %
c) empirically show that our approach\footnote{Our code is available here: \par https://github.com/sophieyueguo/explainable\_action\_advising}. improves policy returns and sample efficiency in single-agent and multi-agent environments, when compared to multiple state-of-the-art baselines.


\section{Related Work}

\textbf{Action Advising (AA)}: Action advising in reinforcement learning is an approach for transferring knowledge from a teacher to a student by giving advice in the form of state-action pairs~\cite{da2019survey}.
Important early work introduced the concept of teaching on a budget, and described heuristic methods for when to give advice in single-agent missions~\cite{torrey2013teaching}.
Since then, additional work has examined different forms of teacher-student relationships defined by who initiates the advice, e.g.  
(a) student-initiated \cite{da2020uncertainty}, (b) teacher-initiated \cite{fachantidis2019learning}, and (c)  jointly-initiated \cite{amir2016interactive}. 
Most of the work focuses on addressing \textit{when} and \textit{which} action to give, among some of which advising is formed as a reinforcement task that the teacher agent takes the student's performance as its reward  \cite{fachantidis2019learning, gf:Lixto, zimmer2014teacher}.
Other works instead build off of previously introduced heuristic methods for deep reinforcement learning \cite{ilhan2021action, da2017simultaneously}.

\textbf{Action advising in MARL}: In many MARL-based action advising approaches, each agent is simultaneously both a student and a teacher and they take alternate teaching roles~\cite{da2017simultaneously, gf:Lixto, kim2019learning}.
The agents are not required to possess expert knowledge a priori, and agents are capable of probabilistically rejecting 
teammates'
advice~\cite{da2017simultaneously}.
To improve efficiency, option learning is utilized to tell which agent's policy is the best to learn \cite{yang2021efficient}, while distillation and value-matching are used to combine knowledge from homogeneous agents to decompose the multi-agent task \cite{wadhwania2019policy}.
Teaming in partially observable environments in decentralized settings~\cite{ilhan2019teaching} has been explored, in which heuristic teaching methods are employed to send advice to a decentralized team.
Conversely, centralized settings have also received attention
~\cite{gupta2021hammer},
where a single centralized teacher observes the entire space and communicates advice to student agents.
None of the above works consider explainability or transparency with respect to the teacher's advice.
While sub-optimal teachers are considered \cite{subramanian2022multi}, there is no mechanism for a student to identify the reasoning of sub-optimality unlike in our approach.

\textbf{Reusing Action Advice}:
Recent work has explored the reuse of advice in the form of state-action pairs so as to maximize communication efficiency.
In one approach~\cite{zhu2020learning}, state-action pairs are directly stored in memory and later selectively re-used.
An imitation learning module has been proposed to imitate the teacher's advice, such that advice can be generated after the budget is exhausted
~\cite{ilhan2021action}.
Prior work has also investigated a teacher capable of providing a qualitative assessment of a given state along with a piece of advice, such that students are aware of how favorable a given state is~\cite{anand2021enhanced}.
However, these approaches all struggle to generalize action advice to novel states and environments.

    
\section{Background}

\textbf{Multi-Agent Reinforcement Learning}: 
We consider the problem of transferring knowledge from a teacher to a student in a 
multi-agent RL setting. The Markov Decision Process (MDP) for this
is a tuple of ($\mathcal{N}, \mathcal{S},\mathcal{A},\mathcal{R},\mathcal{T})$, where $\mathcal{N} = \{0...n\}$ is a set of agents, $\mathcal{S}$ is a set of states, $\mathcal{A} = \Pi_i \mathcal{A}_i$ is a set of joint actions , $\mathcal{R}: \mathcal{S} \mapsto \mathbb{R}$ is a reward function, $\mathcal{T}(s,a_0,a_1...a_n,s') = P(s'|s,a_0,a_1...a_n)$ is a transition function. 
There is a set of policies $\pi_i(s, a_i) = P(a_i|s)$ for each agent $i$, where the single-agent setting is a special case.
The goal is to learn a policy which maximizes the discounted cumulative reward.


\textbf{Decision Trees and Policy Distillation}: %
A precondition of our method is the distillation of a black-box teacher agent policy into an interpretable representation, i.e., a decision tree which is commonly employed as an interpretable proxy model~\cite{gilpin2018explaining}.
Given a pretrained neural network policy, we extract it into an equivalent decision tree model utilizing the distillation method from the \textit{Verifiability via Iterative Policy ExtRaction} (VIPER)~\cite{bastani2018verifiable} framework.
VIPER extracts a suitable decision tree policy $\hat{\pi}^*$ from a policy network $\pi^*$ through an iterative process of sampling trajectories and training a decision tree with standard algorithms, in a {\sc Dagger}-based 
\cite{ross2011reduction}
setup.
Over the course of $N$ iterations, VIPER samples trajectories of state-action pairs $(s, \pi^*(s))$ into a running data set $D_t$. This data set is re-sampled according to $(s, a) \sim P((s, a)) \propto \hat{l}(s) \mathds{1}[(s, a) \in D] $ where $\mathds{1}()$ is the indicator function and the loss is computed by:
\begin{equation}
    \hat{l}_t(s) = V_t^{(\pi^*)}(s) - min_{a \in A}Q_t^{(\pi^*)}(s,a)\nonumber
\end{equation}
After re-sampling to get the current data set $D_t$, a single decision tree policy 
is trained.
After $N$ iterations, the best tree policy $\hat{\pi}^*$ out of all candidate policies is selected.

\section{Methodology}

\begin{figure}\centering
    \vspace{1.5mm}
  \includegraphics[width=.99\linewidth]{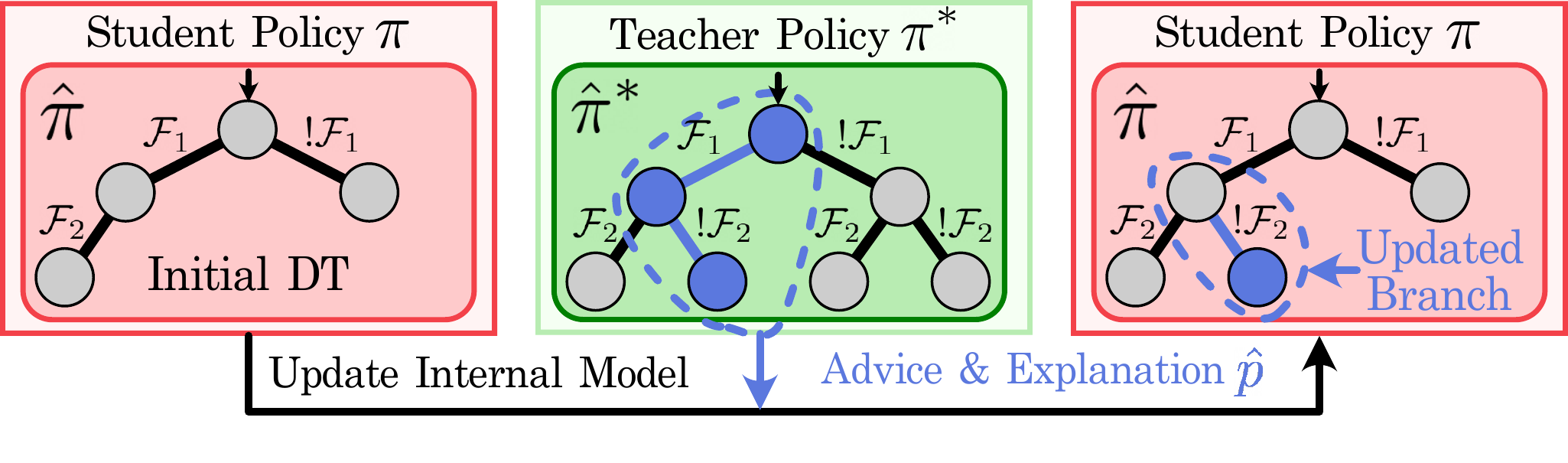}
  \caption{The decision tree reconstruction process that is performed when updating a student's internal model $\hat{\pi}$. The student receives an explanation decision path $\mathcal{F}_1 \land !\mathcal{F}_2$ from the teacher's distilled policy $\hat{\pi}^*$, finds the last common node $\mathcal{F}_1$ currently in its model, and adds all subsequent missing nodes -- in this case only $!\mathcal{F}_2$.
  \vspace{-0.5cm}}
  \label{fig:reconstruction}
\end{figure}



We introduce Explainable Action Advising (see Fig.~\ref{fig:overview}), in which a teacher agent provides action advice as well as an explanation to a student agent.

\subsection{Explainable Action Advising} \label{subsection: EAA}

\begin{algorithm}
\textbf{Input}:  Distilled teacher policies $\hat{\pi}_1^*,... \hat{\pi}_n^*$ for $n$-agent mission, heuristic function $h(\cdot)$ \\
\textbf{Parameter}: Advice budget $b$, advice decay rate $\gamma$\\
\begin{algorithmic}[1]
\STATE Internal model $\hat{\pi}_i = \emptyset$, student $\pi_i = \emptyset$  $\forall{} i = 1, \dots, n$\label{alg:setup}
\FOR{iter $j = 1, 2, \dots$}
\FOR{time step $t = 1, \dots, T$}
\FOR{agent $i = 1, \dots, n$}
\IF{$\gamma^j < \alpha \sim \mathcal{U}[0, 1]$ and $s_i^t \in \hat{\pi}_i$}
\STATE take action $a_i^t = \hat{\pi}_i(s_i^t)$ \label{alg:advice_reuse}
\ELSIF{$b > 0$ and $h(s_i^t)$}
\STATE{give advice $\hat{a}_i^t$ and expl. $\hat{p}_i^t = \texttt{path}(\hat{\pi}_i^*, s_i^t)$} \label{alg:advice_issue}
\STATE{update internal model $\hat{\pi}_i = \texttt{store}(\hat{\pi}_i, \hat{p}_i^t$)}
\STATE{take action $\hat{a}_i^t$ }
\ELSE
\STATE{take action $a_i^t = \pi_i(s_i^t)$}
\label{alg:advice_own}
\ENDIF
\ENDFOR
\ENDFOR
\STATE Update $\pi_i$ via policy optimization for $i = 1, \dots, n$
\ENDFOR
\end{algorithmic}
\caption{Explainable Action Advising}
\label{alg:algorithm}
\end{algorithm}


An outline of the EAA algorithm is shown in Algorithm~\ref{alg:algorithm}.
The distilled teacher policies $\hat{\pi}_1^*, \hat{\pi}_2^*... \hat{\pi}_n^*$ are provided as input to EAA -- one for each student agent -- along with a heuristic function $h(\cdot)$ which determines when advice should be issued to an agent.
We define two hyperparameters: an advising budget $b$ and a decay parameter $\gamma$.
The budget $b$ is used to limit the amount of advice that is issued during the student's training, which is useful for limiting the communication between teacher and student, and motivated by preventing cognitive overload in potential human students ~\cite{torrey2013teaching}.
The decay rate $\gamma$ determines the likelihood with which an agent will re-use advice that has been stored in its internal model $\hat{\pi}_i$, giving the student freedom to execute its own policy after it has sufficiently learned from the teacher's advice and potentially exceed the teacher's own performance.


Our algorithm operates over multiple iterations, during which we roll out a policy and sample $T$ timesteps from the environment.
For each time step $t$, one of three things happens: a) the student will query its internal model for an action, b) the teacher will issue advice and an explanation which the student will incorporate into its internal model and then follow, or c) the student will sample an action from its own current policy.
In the first case on Line~\ref{alg:advice_reuse}, if the student has previously received and stored advice applicable to its current state, it follows the corresponding action with a probability determined by the decay rate $\gamma$.
Intuitively, the agent should become less reliant on the teacher's prior advice over time and increasingly follow its own actions.

In the second case on Line~\ref{alg:advice_issue}, if the student has not previously received relevant advice the teacher agent determines whether advice should be issued to its student, which is controlled by the heuristic function $h(\cdot)$, the remaining budget $b$, and the uncertainty associated with the advised action such that only sufficiently confident predictions are used as advice.
If this occurs, the teacher samples a leaf node action $\hat{a}_i^t$  and its decision path explanation $\hat{p}_i^t$ from its distilled decision tree and sends them to the student.
The student takes this explanation, incorporates it into its own internal model $\hat{\pi}_i$, and follows the advised action.

Lastly, in the third case on Line~\ref{alg:advice_own} the student simply follows its own current policy $\pi_i$, which is itself updated on each iteration with the data collected from the rollouts.
Additionally, there is an optional check in which a teacher agent may only issue advice if: a) the uncertainty of executing the action is sufficiently small, and b) the advice from its internal model matches the action from its expert policy -- useful if the distilled tree is overly noisy.
If the action uncertainty is too high, the teacher may choose to issue only action advice and withold the explanation from the student.
This allows the student to leverage the teacher's learned action probabilities without committing them to memory.


\subsection{Decision Tree Reconstruction}
\label{sec:method_reconstruction}
\begin{figure*}
    \centering
    \vspace{1.5mm}
    \includegraphics[width=0.8\linewidth]{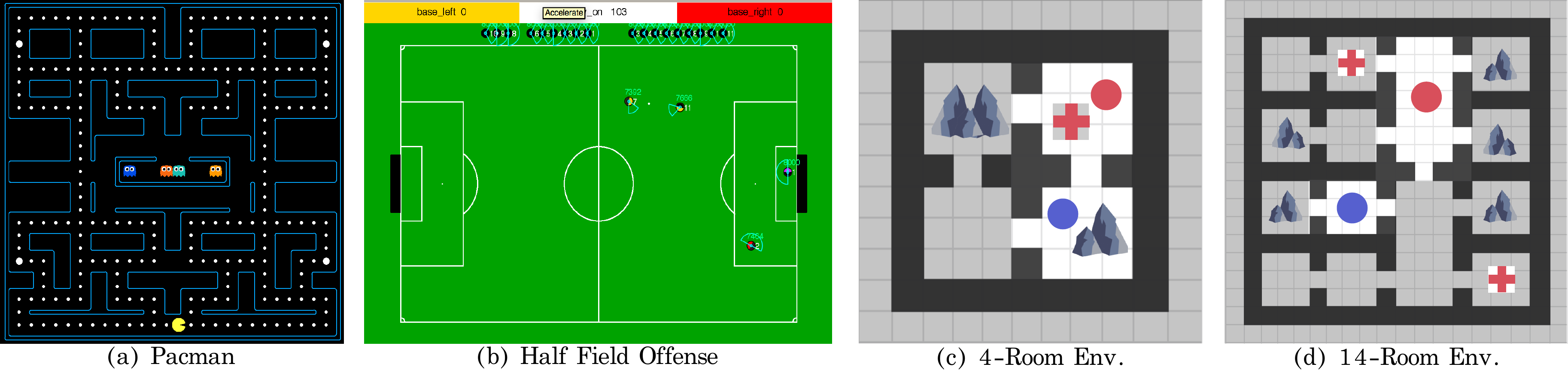}  
    \caption{Environment Screenshots. The 4 and 14 room environment have a medic (red circle), engineer (blue circle) and victims (red plus).
    \vspace{-0.5cm}}
    \label{fig:env_setups}
\end{figure*}
The reconstruction process performed in the \texttt{store} function, shown in Fig.~\ref{fig:reconstruction}, consists of merging the current decision tree $\hat{\pi}_i$ and the decision path explanation, $\hat{p}_i^t$ which reduces to merging two trees.
Following standard decision tree architecture, we assume that the tree consists of a set of $K$ nodes $u_0, \dots, u_K$, where each node contains information regarding its child nodes and branching condition.
By transferring the teacher's tree in a piece-wise fashion, we transmit only useful advice which aids in the generalization process -- uncertain actions are ignored -- while minimizing the amount of information transmitted. The size of the student's tree is conditional on the budget and complexity of the policy.

\subsection{Generalization and Transfer Learning} \label{subsection: Generalization and Transfer Learning}

Using a decision tree representation for the internal model allows the student to generalize the advice to novel states which differ from that in which it was issued.
We do this by treating the decision tree as a classification model, in which the state $s$ is the input and the action $a$ is the target, and extracting a generalizable classification tree during distillation.
Formally, let the set of all possible observed features be $\mathcal{F}_S$ and the set of features modeled by our reconstructed decision tree be $\mathcal{F}_R$ where $\mathcal{F}_R \subseteq \mathcal{F}_S$.
Then a given decision path corresponding to a feature vector $f \in \mathcal{F}_S$ will also generalize to any other feature vector $f' \in \mathcal{F}_S$ as long as the features in which they differ are not contained in $\mathcal{F}_R$.
Such generalization leads to improved advice reuse, and as a result improved learning performance as is empirically shown in Sec.~\ref{subsection: Transfer Learning}.

Similarly, a decision tree representation also lends itself to transfer learning, by allowing a student to recognize situations in which advice is \emph{not} applicable and as such should not be followed, e.g. when the teacher is trained in a different environment.
Let us denote the set of features in the source environment as $\mathcal{F}_S$ and the set of features in the target environment as $\mathcal{F}_T$.
We impose an additional constraint on the student such that it \emph{rejects} advice if the decision path explanation contains features in the set $\mathcal{F}_S - \mathcal{F}_T$ or if the state vector contains features in $\mathcal{F}_T - \mathcal{F}_S$, and the student instead follows its own policy.
Intuitively, we only want to follow advice if it is dependent on features found in both environments.
We refer to this process as \emph{learning through reflection}, and empirically demonstrate that this yields significantly improved policy transfer in Sec.~\ref{subsection: Transfer Learning}.

\section{Experiments}
    


\begin{figure*}
    \centering
    \vspace{1.5mm}
    \includegraphics[width=1\textwidth]{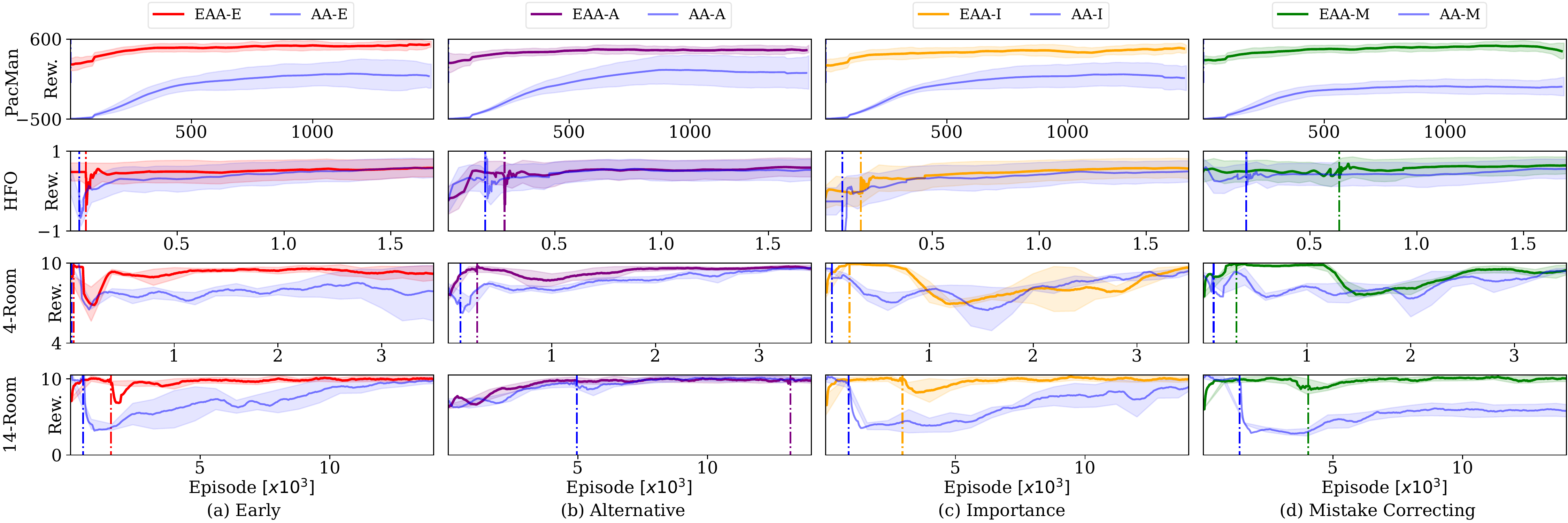}
    \caption{Comparing Explainable Action Advising (Early: Red, Alternative: Purple, Importance: Yellow, Mistake Correcting: Green) with Action Advising (Blue) on the PacMan, HFO, and USAR (four and fourteen room) environments. Dash-Dotted lines indicate when advising budget is exhausted. Results are averaged over 5 independent trials except for 14-room over 3.
    \vspace{-0.5cm}}
    \label{fig:simple_heuristic}
\end{figure*}

We empirically evaluate EAA in three simulated environments:
Pacman, Half Field Offense (HFO), and Urban Search and Rescue (USAR).
In each of the environments we first train a teacher policy to convergence and use this policy to subsequently advise a student policy using EAA as well as several baseline algorithms.
To evaluate EAA's transfer learning ability, we perform an additional experiment in the USAR environment where the teacher is trained in a variant where a feature is missing, and then advise a student in the environment with that feature.
In total, we evaluate EAA against $11$ benchmark methods: four action advising methods in Fig.~\ref{fig:simple_heuristic}, three memory-based action advising methods in Fig.~\ref{fig:simple_benchmark} (a), one adaptive teacher action advising method in Fig.~\ref{fig:simple_benchmark} (b), and three transfer learning techniques in Fig.~\ref{fig:generalization}.

\subsection{Experimental Setup}


Figure~\ref{fig:env_setups} depicts our three environments (Pacman, HFO, and USAR), with the USAR task implemented in a simpler 4-room and more difficult 14-room setup.


\textbf{Pacman}: %
a single-agent scenario with a discrete state space
~\cite{pacman}.
We use the standard single-agent environment~\cite{pacman} with the \textit{OriginalClassic} layout (Fig.~\ref{fig:env_setups} (a)) and $4$ aggressive ghosts.
An additional reward penalty of $-1$ is received when an invalid action is predicted, in place of action masking.
The student and teacher policies were trained via PPO~\cite{schulman2017proximal} for $10,000$ episodes with an advice budget of $b = 10$.
The teacher and distilled teacher achieved average rewards of $450$.

\textbf{HFO}: %
a multi-agent scenario with a continuous state space where two homogeneous RoboCup agents play as a team ~\cite{kalyanakrishnan2006half, hausknecht2016half}.
We use the standard 2v2 environment~\cite{kalyanakrishnan2006half} (Fig.~\ref{fig:env_setups} (b)) and train teacher and student policies for the offense agents with SARSA~\cite{sutton2018reinforcement} as in prior work~\cite{da2017simultaneously} for $2,000$ episodes with an advice budget of $300$.
The teacher and distilled teacher achieved an average reward of $0.8$.

\textbf{USAR}: %
a multi-agent scenario with a discrete state space and heterogeneous agents with different action spaces and hard coordination constraints ~\cite{lewis2019developing, guo2021transfer, freeman2021}.
We have two layouts: a simple 4-room layout (Fig.~\ref{fig:env_setups} (c)) and a more complex 14-room layout (Fig.~\ref{fig:env_setups} (d)).
The environment consists of rubble which can be cleared and victims which must be healed, and a team of two agents: a medic who heals unblocked victims and an engineer who can clear rubble from blocked victims.
The reward is team-level and gives $+10$ for each victim healed.
Both layouts contain rubble in every room, while the 4-room layout contains 1 victim randomly distributed and the 14-room layout contains 2 victims.
The teacher and student policies are trained using COMA~\cite{foerster2018counterfactual} for $16,000$ episodes with an advice budget of $100,000$.
The teacher and distilled teacher achieved an average reward of $10$ in both layouts, indicating a sub-optimal teacher for the 14-room layout.
This task necessitates cooperation between a medic and an engineer robot to save trapped victims -- the engineer must remove rubble before the medic comes across and heals the victim, and is the most difficult environment. 

\subsection{Benchmarks}\label{Benchmarks}


In all experiments, we define four variants of EAA utilizing the heuristic functions established in prior work~\cite{torrey2013teaching} as $h(\cdot)$: EAA with early advising (EAA-E), alternative advising (EAA-A), importance advising (EAA-I), and mistake correcting advising (EAA-M).
EAA-Median indicates the median performance of all EAA variants.
The heuristics are the same as in the Action Advising baselines and are defined below.

\textbf{Action Advising}: %
We compare EAA to four heuristic-based Action Advising algorithms~\cite{torrey2013teaching} in all environments:
\begin{enumerate}
    \item \textit{
    (AA-E) Early
    }: Continuously advise at each state.
    \item \textit{
    (AA-A) Alternative
    }: Advise with a fixed frequency.
    \item \textit{
    (AA-I) Importance
    }: Advise when the state importance $I(s) > \text{threshold}$.
    \item \textit{
    (AA-M) Mistake Correcting
    }: Advise if $I(s) > \text{threshold}$ and $a^{student} \neq a^*$.
\end{enumerate}

\begin{figure}
  \centering
  \includegraphics[width=.99\linewidth]{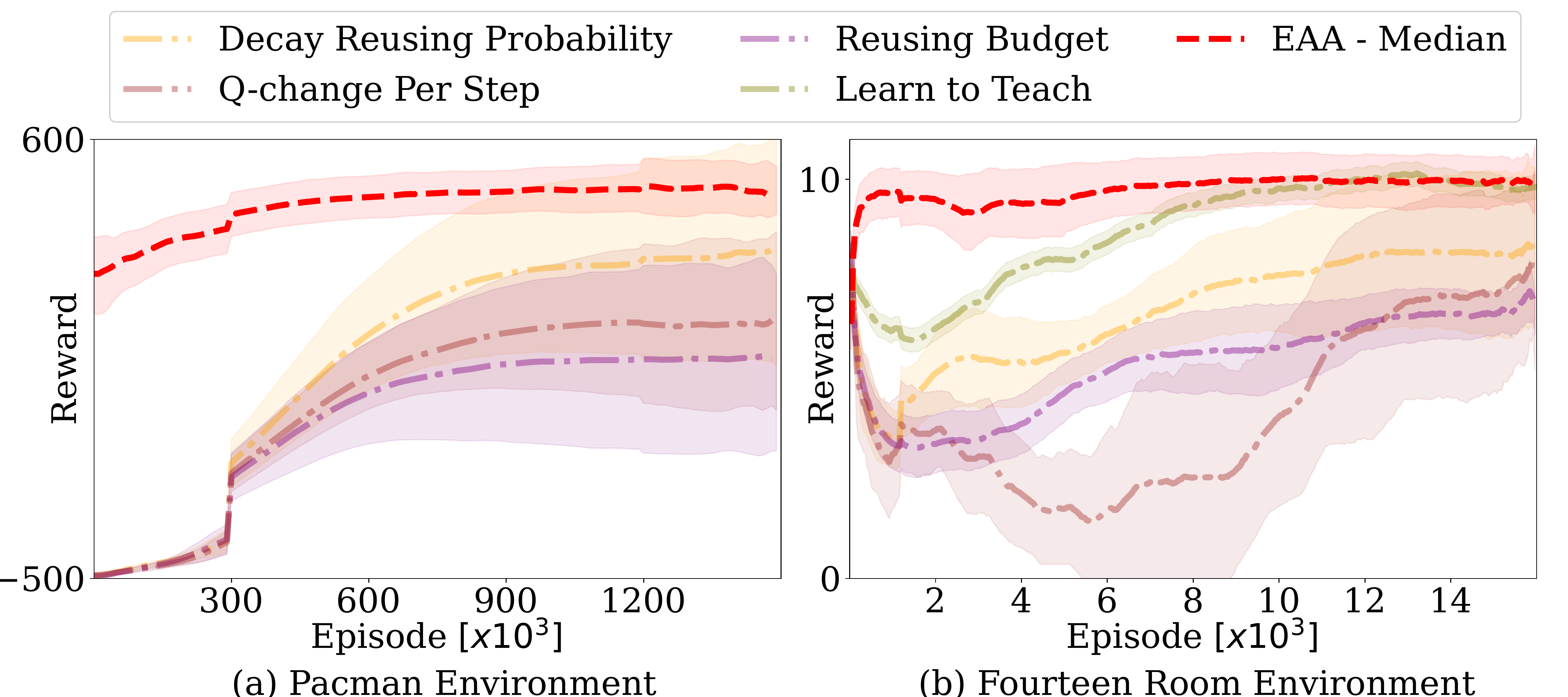}
  \caption{Baseline of Pacman (left) and 14-room USAR comparing EAA to SOTA algorithms. Results are averaged over 3 random seeds except for Learn to Teach which uses 1 due to run-time.
  \vspace{-0.5cm}}
  \label{fig:simple_benchmark}
\end{figure}

\textbf{Memory-based Action Advising}: %
For the discrete state environments, we compare EAA to three recently introduced~\cite{zhu2020learning} benchmark algorithms: Q-change Per Step, Decay Reusing Probability, and Reusing Budget.
These works are conceptually similar to ours in that students have a memory and reuse teacher advice in the form of state-action pairs, so they serve as a strong state-of-the-art comparison.

\textbf{Action Advising with Adaptive Teacher}: %
In the USAR environment, we further compare EAA against an approach in which reinforcement learning is used to learn an optimal teacher~\cite{zimmer2014teacher, gf:Lixto},  and denote this as \emph{Learn to Teach}.
This algorithm learns over multiple sessions, within each of which the teacher learns to allocate its advice budget with the constraint that the student's policy must converge.
All methods except for Learn to Teach are trained for $5$ random seeds.
Due to the computational demand of the inner reinforcement learning loop, Learn to Teach (100x slower) is only evaluated once for USAR.

\begin{figure}
  \centering
  \includegraphics[width=.99\linewidth]{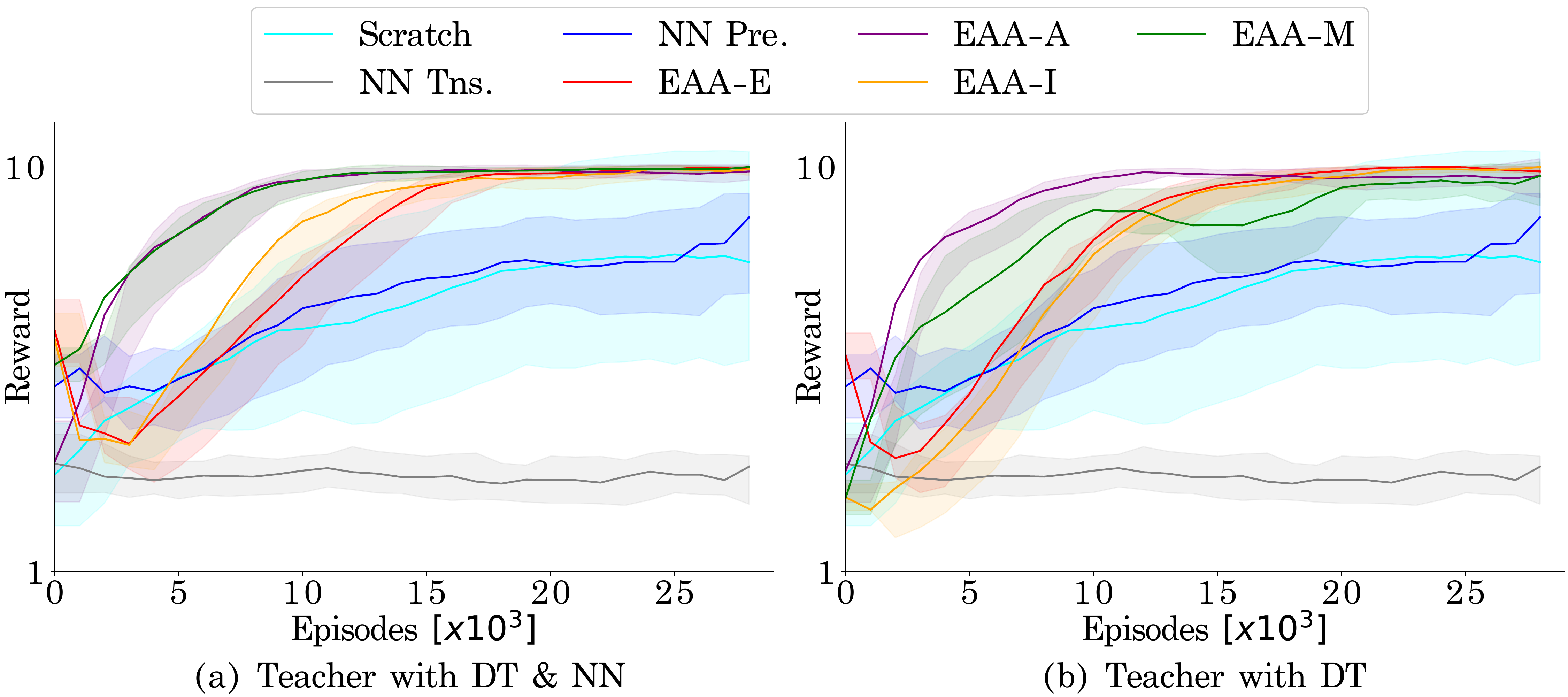}
  \label{fig:generalization_dt}
\caption{Transfer learning performance for students training in the USAR environment with rubble by a teacher who was originally trained without rubble.
The teacher only has access to the extracted decision tree where the teacher NN is not present in (b), compared to (a) where the teacher NN is present. Results are averaged over 5 independent trials.
\vspace{-0.5cm}}
\label{fig:generalization}
\end{figure}

\begin{figure*}
    \centering
    \vspace{1.5mm}
    \includegraphics[width=1\textwidth]{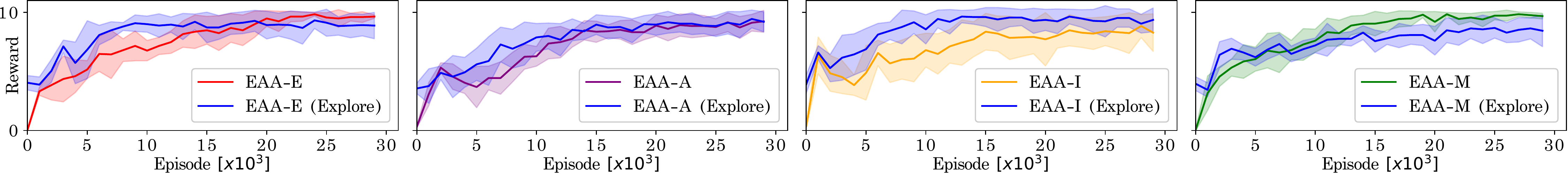}
    \caption{A comparison of EAA with reflection, 
    called EAA (Explore),
    to standard EAA. The student is pretrained in a source environment (4-room no rubble) and transferred to a target environment (4-room with rubble). Results are averaged over 5 independent trials. 
    \vspace{-0.5cm}}
    \label{fig:transfer}
\end{figure*}

\textbf{Fine-tuning}: %
In the transfer learning experiments, we compare EAA against the standard transfer learning technique in which a policy is pre-trained in a source environment, and then fine-tuned in a target environment, denoted as NN Pre.
We also include results for the pre-trained policy without fine-tuning (NN Tns.) and a policy baseline trained from scratch (Scratch).
\subsection{Results: Advice Re-use}
\label{sec:results_advice_reuse}

Figures~\ref{fig:simple_heuristic} and \ref{fig:simple_benchmark} show the results of our EAA variants to the action advising baselines described above, and are intended to demonstrate improved policy return and sample efficiency owing to advise re-use from explanations.









\textbf{Comparison to Action Advising}: %
In Fig.~\ref{fig:simple_heuristic}, we compare our four EAA variants described in Sec.~\ref{Benchmarks} to AA~\cite{torrey2013teaching} in all environments.
We observe that EAA converges to the teacher's reward significantly faster than the AA baselines and with lower reward variance, indicating more training stability.
Of note is PacMan, where the lack of initial state randomization enables heavy advice re-use and offsets the small advice budget, allowing EAA to perform remarkably well.
In HFO, EAA achieves only a slight improvement over AA; we conjecture that this is due to the simplicity of the task -- it was designed for classical methods such as SARSA, which enables AA to perform well.
Advise re-use and generalization is evident in the HFO and USAR environments where EAA results in later budget exhaustion (dashed lines, Fig.~\ref{fig:simple_heuristic}) across all heuristics.
This is not evident in PacMan simply due to the extremely small advice budget; it is spent quickly at the start of training.
All algorithms except for EAA-A (Fig.~\ref{fig:simple_heuristic} (b)) also suffer from a large drop in reward when the teacher's budget is exhausted.
While this is expected, it is interesting to observe that our EAA variants tend to recover quicker from this drop than AA.

\textbf{Comparison to Memory-based Action Advising}:
In Fig.~\ref{fig:simple_benchmark}, we see that EAA again demonstrates both improved convergence rate and maximal policy return when compared to the memory-based action advising baselines, achieving optimal performance in both the Pacman and USAR 14 room environments.
Q-change Per Step displays the worst performance because it assumes a strict increase in Q-values when the student receives advice, otherwise discarding it.
However, this assumption is often violated, especially in USAR where an action such as removing rubble does not directly result in an increase in Q-values -- only healing victims does.
Decay Reusing Probability and Reusing Budget are able to achieve optimal performance, yet take significantly longer to converge than EAA. 

\textbf{Comparison to Action Advising with Adaptive Teacher}: %
%
Figure~\ref{fig:simple_benchmark} (b) shows the comparison the adaptive teacher action advising method, Learn to Teach, in the USAR 14-room layout.
While Learn to Teach converges to the teacher's reward, it takes considerably longer to do so, requiring nearly $12,000$ episodes compared to EAA's $6,000$.


\subsection{Results: Transfer Learning}

Figures~\ref{fig:generalization} and \ref{fig:transfer} show the results when self-reflection is used to selectively transfer advice from a teacher trained in a different environment than the student.

\textbf{Comparison to Fine-tuning}: \label{subsection: Transfer Learning}%
Figure~\ref{fig:generalization} shows EAA's performance when transferring advice from a source environment to a target environment, as compared to standard transfer learning approaches.
The teacher is a trained expert in a USAR 4-room source environment \textit{without} rubble, and is used to advise a student in a target environment \emph{with} rubble, in which it is a non-expert due to the additional coordination requirement of removing it.
As discussed in Sec.~\ref{subsection: Generalization and Transfer Learning}, the student rejects advice if the teacher's advised explanation contains no features related to rubble when encountering a room with rubble.
Fig.~\ref{fig:generalization} (a) EAA has access to both the distilled and original policy and does not store explanations in which the action has a low probability, as discussed in Sec.~\ref{subsection: EAA}.
This is the default behavior used for prior experiments in Sec.~\ref{sec:results_advice_reuse}, and represents the best-case transfer learning scenario.
In Fig.~\ref{fig:generalization} (b), the EAA variants are trained by a teacher who only has access to its distilled decision tree and not its original policy.
This represents a more difficult task, and establishes a baseline for the later case where there is no teacher and the student relies only on its internal model trained in the source environment (Fig.~\ref{fig:transfer}).

In Figure~\ref{fig:generalization} all four variants of EAA significantly outperform the neural network baselines and converge to the optimal reward within $10,000$ episodes.
The transferred teacher policy (NN Tns.) serves as a baseline of the non-expert teacher behavior in the target environment without any additional training; this is the upper limit on the expected sub-optimal advice performance.
Notably neither training from scratch (Scratch) nor pre-training on the source environment followed by fine-tuning (NN pre.) consistently converges to the optimal reward.
Given that EAA performs nearly as well in Fig.~\ref{fig:generalization} (b) as Fig.~\ref{fig:generalization} (a), we conclude that maintaining the original black-box teacher policy is unnecessary.

\textbf{Transfer Learning through Self-Reflection}: %
Lastly, Fig.~\ref{fig:transfer} shows the results when a student trained via EAA in a source environment continues training in a target environment, when no teacher -- and therefore no advising -- is present.
We implement a version of EAA with reflection in which the student examines its own reconstructed decision tree and ignores actions learned from unhelpful advice and instead executes an exploratory action (EAA Explore).
When compared to an ablated student without reflection, we find that reflection not only leads to improved rewards at the start of training, but also an improved convergence rate for EAA-E, EAA-A, and EAA-I variants.
These results show that not only does EAA improve policy transfer to novel environments in the presence of a teacher, but it even yields benefits when the student operates in novel environments \emph{by itself}.


\section{Conclusion}
We propose EAA,
a novel action advising algorithm in which a teacher agent issues explanations in addition to action advice to student agents.
Our results show that EAA improves both policy reward as well as convergence speed 
compared to eight state-of-the-art benchmark algorithms, while also improving advice generalization and transfer to novel environments through
reflection.
We demonstrate that our method is especially effective in budget-constrained advising, as each piece of advice now conveys more information and helps reduce communication frequency with respect to advice in the case of artificial agents.
We designed EAA with human students in mind, and intend to investigate the degree to which our explanations are intelligible to humans in the future.



\addtolength{\textheight}{-5.9cm}


\addtolength{\textheight}{2cm}

\bibliographystyle{IEEEtran}
\bibliography{IEEEabrv,reference}

\begin{thebibliography}{10}
\providecommand{\url}[1]{#1}
\csname url@samestyle\endcsname
\providecommand{\newblock}{\relax}
\providecommand{\bibinfo}[2]{#2}
\providecommand{\BIBentrySTDinterwordspacing}{\spaceskip=0pt\relax}
\providecommand{\BIBentryALTinterwordstretchfactor}{4}
\providecommand{\BIBentryALTinterwordspacing}{\spaceskip=\fontdimen2\font plus
\BIBentryALTinterwordstretchfactor\fontdimen3\font minus
  \fontdimen4\font\relax}
\providecommand{\BIBforeignlanguage}[2]{{%
\expandafter\ifx\csname l@#1\endcsname\relax
\typeout{** WARNING: IEEEtran.bst: No hyphenation pattern has been}%
\typeout{** loaded for the language `#1'. Using the pattern for}%
\typeout{** the default language instead.}%
\else
\language=\csname l@#1\endcsname
\fi
#2}}
\providecommand{\BIBdecl}{\relax}
\BIBdecl

\bibitem{da2020agents}
F.~L. Da~Silva, G.~Warnell, A.~H.~R. Costa, and P.~Stone, ``Agents teaching
  agents: a survey on inter-agent transfer learning,'' \emph{Autonomous Agents
  and Multi-Agent Systems}, vol.~34, no.~1, pp. 1--17, 2020.

\bibitem{torrey2013teaching}
L.~Torrey and M.~Taylor, ``Teaching on a budget: Agents advising agents in
  reinforcement learning,'' in \emph{Proceedings of the 2013 international
  conference on Autonomous agents and multi-agent systems}, 2013, pp.
  1053--1060.

\bibitem{gf:Lixto}
S.~Omidshafiei, D.-K. Kim, M.~Liu, G.~Tesauro, M.~Riemer, C.~Amato,
  M.~Campbell, and J.~P. How, ``Learning to teach in cooperative multiagent
  reinforcement learning,'' in \emph{Proceedings of the AAAI Conference on
  Artificial Intelligence}, vol.~33, no.~01, 2019, pp. 6128--6136.

\bibitem{zhu2020learning}
C.~Zhu, Y.~Cai, H.-f. Leung, and S.~Hu, ``Learning by reusing previous advice
  in teacher-student paradigm,'' in \emph{Proceedings of the 19th International
  Conference on Autonomous Agents and MultiAgent Systems}, 2020, pp.
  1674--1682.

\bibitem{da2020uncertainty}
F.~L. Da~Silva, P.~Hernandez-Leal, B.~Kartal, and M.~E. Taylor,
  ``Uncertainty-aware action advising for deep reinforcement learning agents,''
  in \emph{Proceedings of the AAAI Conference on Artificial Intelligence},
  vol.~34, no.~04, 2020, pp. 5792--5799.

\bibitem{ho2016generative}
J.~Ho and S.~Ermon, ``Generative adversarial imitation learning,''
  \emph{Advances in neural information processing systems}, vol.~29, 2016.

\bibitem{grover2022semantically}
J.~Grover, Y.~Lyu, W.~Luo, C.~Liu, J.~Dolan, and K.~Sycara,
  ``Semantically-aware pedestrian intent prediction with barrier functions and
  mixed-integer quadratic programming,'' \emph{IFAC-PapersOnLine}, vol.~55,
  no.~41, pp. 167--174, 2022.

\bibitem{gao2018reinforcement}
Y.~Gao, H.~Xu, J.~Lin, F.~Yu, S.~Levine, and T.~Darrell, ``Reinforcement
  learning from imperfect demonstrations,'' \emph{arXiv preprint
  arXiv:1802.05313}, 2018.

\bibitem{helyer2015learning}
R.~Helyer, ``Learning through reflection: the critical role of reflection in
  work-based learning (wbl),'' \emph{Journal of Work-Applied Management}, 2015.

\bibitem{chi1994eliciting}
M.~T. Chi, N.~De~Leeuw, M.-H. Chiu, and C.~LaVancher, ``Eliciting
  self-explanations improves understanding,'' \emph{Cognitive science},
  vol.~18, no.~3, pp. 439--477, 1994.

\bibitem{da2019survey}
F.~L. Da~Silva and A.~H.~R. Costa, ``A survey on transfer learning for
  multiagent reinforcement learning systems,'' \emph{Journal of Artificial
  Intelligence Research}, vol.~64, pp. 645--703, 2019.

\bibitem{fachantidis2019learning}
A.~Fachantidis, M.~E. Taylor, and I.~Vlahavas, ``Learning to teach
  reinforcement learning agents,'' \emph{Machine Learning and Knowledge
  Extraction}, vol.~1, no.~1, pp. 21--42, 2019.

\bibitem{amir2016interactive}
O.~Amir, E.~Kamar, A.~Kolobov, and B.~Grosz, ``Interactive teaching strategies
  for agent training,'' 2016.

\bibitem{zimmer2014teacher}
M.~Zimmer, P.~Viappiani, and P.~Weng, ``Teacher-student framework: a
  reinforcement learning approach,'' in \emph{AAMAS Workshop Autonomous Robots
  and Multirobot Systems}, 2014.

\bibitem{ilhan2021action}
E.~Ilhan, J.~Gow, and D.~Perez-Liebana, ``Action advising with advice imitation
  in deep reinforcement learning,'' \emph{Proceedings of the AAMAS Conference},
  pp. 629--637, 2021.

\bibitem{da2017simultaneously}
F.~L. Da~Silva, R.~Glatt, and A.~H.~R. Costa, ``Simultaneously learning and
  advising in multiagent reinforcement learning,'' in \emph{Proceedings of the
  16th conference on autonomous agents and multiagent systems}, 2017, pp.
  1100--1108.

\bibitem{kim2019learning}
D.-K. Kim, M.~Liu, S.~Omidshafiei, S.~Lopez-Cot, M.~Riemer, G.~Habibi,
  G.~Tesauro, S.~Mourad, M.~Campbell, and J.~P. How, ``Learning hierarchical
  teaching policies for cooperative agents,'' \emph{arXiv preprint
  arXiv:1903.03216}, 2019.

\bibitem{yang2021efficient}
T.~Yang, W.~Wang, H.~Tang, J.~Hao, Z.~Meng, H.~Mao, D.~Li, W.~Liu, Y.~Chen,
  Y.~Hu \emph{et~al.}, ``An efficient transfer learning framework for
  multiagent reinforcement learning,'' \emph{Advances in Neural Information
  Processing Systems}, vol.~34, pp. 17\,037--17\,048, 2021.

\bibitem{wadhwania2019policy}
S.~Wadhwania, D.-K. Kim, S.~Omidshafiei, and J.~P. How, ``Policy distillation
  and value matching in multiagent reinforcement learning,'' in \emph{2019
  IEEE/RSJ International Conference on Intelligent Robots and Systems
  (IROS)}.\hskip 1em plus 0.5em minus 0.4em\relax IEEE, 2019, pp. 8193--8200.

\bibitem{ilhan2019teaching}
E.~Ilhan, J.~Gow, and D.~Perez-Liebana, ``Teaching on a budget in multi-agent
  deep reinforcement learning,'' in \emph{2019 IEEE Conference on Games
  (CoG)}.\hskip 1em plus 0.5em minus 0.4em\relax IEEE, 2019, pp. 1--8.

\bibitem{gupta2021hammer}
N.~Gupta, G.~Srinivasaraghavan, S.~K. Mohalik, and M.~E. Taylor, ``Hammer:
  Multi-level coordination of reinforcement learning agents via learned
  messaging,'' \emph{arXiv preprint arXiv:2102.00824}, 2021.

\bibitem{subramanian2022multi}
S.~G. Subramanian, M.~E. Taylor, K.~Larson, and M.~Crowley, ``Multi-agent
  advisor {Q}-learning,'' \emph{Journal of Artificial Intelligence Research},
  vol.~74, pp. 1--74, 2022.

\bibitem{anand2021enhanced}
D.~Anand, V.~Gupta, P.~Paruchuri, and B.~Ravindran, ``An enhanced advising
  model in teacher-student framework using state categorization,'' in
  \emph{Proceedings of the AAAI Conference on Artificial Intelligence},
  vol.~35, no.~8, 2021, pp. 6653--6660.

\bibitem{gilpin2018explaining}
L.~H. Gilpin, D.~Bau, B.~Z. Yuan, A.~Bajwa, M.~Specter, and L.~Kagal,
  ``Explaining explanations: An overview of interpretability of machine
  learning,'' in \emph{2018 IEEE 5th International Conference on data science
  and advanced analytics (DSAA)}.\hskip 1em plus 0.5em minus 0.4em\relax IEEE,
  2018, pp. 80--89.

\bibitem{bastani2018verifiable}
O.~Bastani, Y.~Pu, and A.~Solar-Lezama, ``Verifiable reinforcement learning via
  policy extraction,'' \emph{arXiv preprint arXiv:1805.08328}, 2018.

\bibitem{ross2011reduction}
S.~Ross, G.~Gordon, and D.~Bagnell, ``A reduction of imitation learning and
  structured prediction to no-regret online learning,'' in \emph{Proceedings of
  the fourteenth international conference on artificial intelligence and
  statistics}.\hskip 1em plus 0.5em minus 0.4em\relax JMLR Workshop and
  Conference Proceedings, 2011, pp. 627--635.

\bibitem{pacman}
{John DeNero, Dan Klein, Ketrina Yim, and Pieter Abbeel}, ``U{C} {B}erkeley
  {CS}188 intro to {AI},'' University of California, Berkeley.

\bibitem{schulman2017proximal}
J.~Schulman, F.~Wolski, P.~Dhariwal, A.~Radford, and O.~Klimov, ``Proximal
  policy optimization algorithms,'' \emph{arXiv preprint arXiv:1707.06347},
  2017.

\bibitem{kalyanakrishnan2006half}
S.~Kalyanakrishnan, Y.~Liu, and P.~Stone, ``Half field offense in robocup
  soccer: A multiagent reinforcement learning case study,'' in \emph{Robot
  soccer world cup}.\hskip 1em plus 0.5em minus 0.4em\relax Springer, 2006, pp.
  72--85.

\bibitem{hausknecht2016half}
M.~Hausknecht, P.~Mupparaju, S.~Subramanian, S.~Kalyanakrishnan, and P.~Stone,
  ``Half field offense: An environment for multiagent learning and ad hoc
  teamwork,'' in \emph{AAMAS Adaptive Learning Agents (ALA) Workshop},
  vol.~3.\hskip 1em plus 0.5em minus 0.4em\relax sn, 2016.

\bibitem{sutton2018reinforcement}
R.~S. Sutton and A.~G. Barto, \emph{Reinforcement learning: An
  introduction}.\hskip 1em plus 0.5em minus 0.4em\relax MIT press, 2018.

\bibitem{lewis2019developing}
M.~Lewis, K.~Sycara, and I.~Nourbakhsh, ``Developing a testbed for studying
  human-robot interaction in urban search and rescue,'' in \emph{Human-Centered
  Computing}.\hskip 1em plus 0.5em minus 0.4em\relax CRC Press, 2019, pp.
  270--274.

\bibitem{guo2021transfer}
Y.~Guo, R.~Jena, D.~Hughes, M.~Lewis, and K.~Sycara, ``Transfer learning for
  human navigation and triage strategies prediction in a simulated urban search
  and rescue task,'' in \emph{2021 30th IEEE International Conference on Robot
  \& Human Interactive Communication (RO-MAN)}.\hskip 1em plus 0.5em minus
  0.4em\relax IEEE, 2021, pp. 784--791.

\bibitem{freeman2021}
J.~T. Freeman, L.~Huang, M.~Woods, and S.~J. Cauffman, ``Evaluating artificial
  social intelligence in an urban search and rescue task environment,'' in
  \emph{AAAI 2021 Fall Symposium, 4-6 Nov 2021}.\hskip 1em plus 0.5em minus
  0.4em\relax AAAI, 2021.

\bibitem{foerster2018counterfactual}
J.~Foerster, G.~Farquhar, T.~Afouras, N.~Nardelli, and S.~Whiteson,
  ``Counterfactual multi-agent policy gradients,'' in \emph{Proceedings of the
  AAAI Conference on Artificial Intelligence}, vol.~32, no.~1, 2018.

\end{thebibliography}





\end{document}